\def\year{2021}\relax
\documentclass[letterpaper]{article} 
\usepackage{aaai21}  
\usepackage{times}  
\usepackage{helvet} 
\usepackage{courier}  
\usepackage[hyphens]{url}  
\usepackage{graphicx} 
\usepackage{subcaption}
\usepackage{natbib}  
\usepackage{caption} 
\usepackage{latexsym}
\usepackage{amsmath}

\newcommand{\ccirc}{\kern0.5ex\vcenter{\hbox{$\scriptstyle\circ$}}\kern0.5ex}

\usepackage{tabularx,booktabs}
\usepackage{multirow}
\usepackage{algorithm,algorithmic}
\usepackage{bbm}
\usepackage[switch]{lineno}
\newcolumntype{C}{>{\centering\arraybackslash}X}
\usepackage{enumerate}

\title{KgPLM: Knowledge-guided Language Model Pre-training via Generative and Discriminative Learning}

\author{Bin He, Xin Jiang, Jinghui Xiao, Qun Liu\\   
}
\affiliations{
	Huawei Noah’s Ark Lab\\
	\{hebin.nlp, Jiang.Xin, xiaojinghui4, qun.liu\}@huawei.com 
}

\begin{document}
	\maketitle

\begin{abstract}
Recent studies on pre-trained language models have demonstrated their ability to capture factual knowledge and applications in knowledge-aware downstream tasks. In this work, we present a language model pre-training framework guided by factual knowledge completion and verification, and use the \textit{generative} and \textit{discriminative} approaches cooperatively to learn the model. Particularly, we investigate two learning schemes, named \textit{two-tower} scheme and \textit{pipeline} scheme, in training the generator and discriminator with shared parameter. Experimental results on LAMA, a set of zero-shot cloze-style question answering tasks, show that our model contains richer factual knowledge than the conventional pre-trained language models. Furthermore, when fine-tuned and evaluated on the MRQA shared tasks which consists of several machine reading comprehension datasets, our model achieves the state-of-the-art performance, and gains large improvements on NewsQA (+1.26 F1) and TriviaQA (+1.56 F1) over RoBERTa.
\end{abstract}

\section{Introduction}

Pre-trained language models such as BERT or RoBERTa learn contextualized word representations on large-scale text corpus through self-supervised learning, and obtain new state-of-the-art results on many downstream NLP tasks \cite{peters2018deep,radford2018improving,devlin2019bert,liu2019roberta}.
Recently, researchers have observed that pre-trained language models can internalize real-word knowledge into their model parameters~\cite{petroni2019language,logan2019barack,talmor2019olmpics}. For example, pre-trained language models are able to answer the questions such as ``\textit{the sky is $\rule{0.5cm}{0.15mm}$}'' or ``\textit{Beethoven was born in $\rule{0.5cm}{0.15mm}$}'' with moderate accuracy. To further explore their potential, researchers have proposed various approaches to guide the pre-training of the language models by injecting different forms of knowledge into them, such as structured knowledge graph or linguistic knowledge  ~\cite{zhang2019ernie,peters2019knowledge,xiong2019pretrained,wang2020k,roberts2020much,joshi2020contextualized}.

\begin{figure}[!t]
	\centering
	\includegraphics[trim={8.5in 0in 0in 5in}, clip, width=\linewidth]{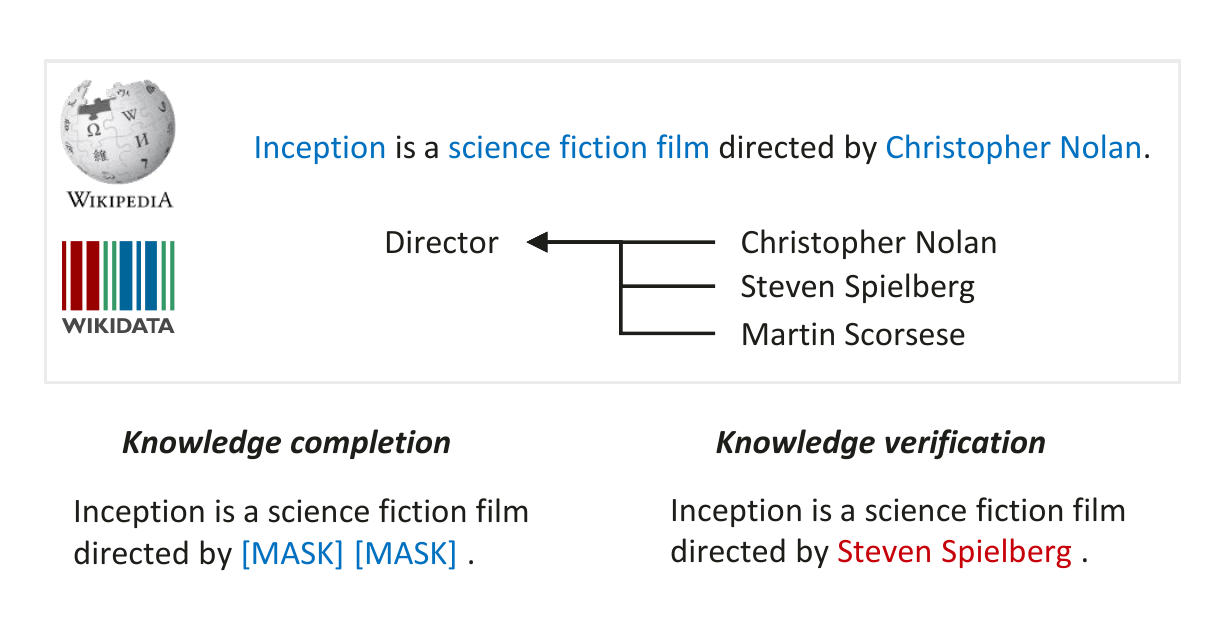}
	\caption{\label{fig:example} An example of \textit{generative} knowledge completion and \textit{discriminative} knowledge verification tasks. The span \textit{Christopher Nolan} is masked for prediction (left) and a ``fake'' span \textit{Steven Spielberg} replaces the original span for detection (right).}
\end{figure}

\begin{table*}[t]
	\centering
	\small
	\begin{tabularx}{\linewidth}{p{5cm}XX}
		\hline
		\textbf{Model} & \textbf{Generative tasks} & \textbf{Discriminative tasks}\\
		\hline
		ERNIE-Baidu~\scriptsize{\cite{sun2019ernie}} & phrase and entity masking & -  \\
		SenseBERT~\scriptsize{\cite{levine2019sensebert}} & supersense prediction & -  \\
		BERT\_CS~\scriptsize{\cite{ye2019align}} & - & multi-choice question answering \\
		LIBERT~\scriptsize{\cite{lauscher2019informing}} & - & lexical relation prediction \\
		LIMIT-BERT~\scriptsize{\cite{zhou2019limit}}  & semantic/syntactic phrase masking & -  \\
		KEPLER~\scriptsize{\cite{wang2019kepler}} & - & knowledge representation learning  \\
		SpanBERT~\scriptsize{\cite{joshi2020spanbert}} & span masking & -  \\
		WKLM~\scriptsize{\cite{xiong2019pretrained}} &  - &  entity replacement checking \\
		K-Adapter~\scriptsize{\cite{wang2020k}} & - & relation classification, dependency relation prediction \\
		T5+SSM~\scriptsize{\cite{roberts2020much}} & salient span masking & - \\
		TEK~\scriptsize{\cite{joshi2020contextualized}} & span masking on TEK-augmented text & - \\
		CN-ADAPT~\scriptsize{\cite{lauscher2020common}} & MLM training on synthetic knowledge corpus & - \\
		\hline
		KgPLM (ours) & knowledge span masking & knowledge span replacement checking \\
		\hline
	\end{tabularx}
	\caption{Comparison of training methods between different knowledge-guided pre-training language models.}\label{tab:knowledge}
\end{table*}

Table~\ref{tab:knowledge} lists some of the previous knowledge-guided pre-trained language models with their training methods. We group them into two categories: \textit{generative} tasks and \textit{discriminative} tasks. Generative tasks are often formulated as predicting the masked tokens given the context. By particularly masking out the words that contain certain types of knowledge (e.g., entities) in generative pre-training, the model can be more adept in memorizing and completing such knowledge. While discriminative tasks are often formulated as a classification problem with respect to the sentence or the tokens. By training on the positive and negative examples constructed according to the external knowledge, the discriminator can be more capable of verifying the true or false knowledge in natural language. Existing research has demonstrated that \textit{generative} and \textit{discriminative} training have their advantages: the former has a large negative sample space so that the model can learn fine-grained knowledge, while the latter avoids the ``\texttt{[MASK]}'' tokens in pre-training, and is therefore more consistent with fine-tuning. On the other hand, generative and discriminative capture the different aspects of data distribution and could be complementary to each other in knowledge consolidation. However, to the best of our knowledge, there is not previous work in combining the two approaches in a systematic way. Inspired by the recent success on the generative-discriminative pre-trained model named ELECTRA~\cite{clark2020electra}, we propose to learn the generator and discriminator jointly in the knowledge-guided pre-training, which we call the KgPLM model.

In this paper, we design \textit{masked span prediction} as the generative knowledge completion task, and \textit{span replacement checking} as the discriminative knowledge verification task. Hybrid knowledge, including link structure of Wikipedia and structured knowledge graph in Wikidata, is used to guide the both tasks. The spans covering the factual knowledge are more likely to be selected for masking or replacement, and the choices of their replacements are also related to the proximity to the original span in the knowledge space. Figure~\ref{fig:example} shows an example of the span masking and replacement tasks. To further explore effective ways to the joint training of the two tasks, we design two learning schemes, which we called \textit{two-tower} scheme and \textit{pipeline} scheme. Basically, the generator and discriminator are trained in parallel with the shared parameters in the two-tower scheme. While in the pipeline scheme, the output of generator is input to the successive discriminative training. The generator and discriminator in our KgPLM model are both pre-trained based on RoBERTa$_{\text{BASE}}$~\cite{liu2019roberta}. They have some additional benefits: 1) the model can be readily extended to much larger pre-training corpus, which keeps some potential room for further improvement; 2) the model retains the same amount of parameters as RoBERTa$_{\text{BASE}}$, and does not require any modifications in fine-tuning for the downstream tasks.

We evaluate the model performance on LAMA~\citep{petroni2019language}, which consists of several zero-shot knowledge completion tasks, and MRQA shared tasks~\citep{fisch2019mrqa}, which include several benchmark question answering datasets. The experiments show the proposed KgPLM, especially that trained with the pipeline scheme, achieves the state-of-the-art performance, and significantly outperform several strong baselines (RoBERTa$_{\text{BASE}}$, SpanBERT$_{\text{BASE}}$~\cite{joshi2020spanbert}, WKLM~\cite{xiong2019pretrained}) on some of the tasks. The results indicate that the knowledge-guided generative and discriminative pre-training provides an effective way to incorporate external knowledge and achieve competitive performance on the knowledge intensive NLP tasks.

\section{Related Work}

Most knowledge-guided pre-training methods can be categorized into groups according to their pre-training objectives
(1) \textit{generative} objectives,
and (2) \textit{discriminative} objectives.

For masked language models, different masking mechanisms are always used to design the generative objectives.
ERNIE-Baidu~\cite{sun2019ernie} introduces new masking units such as phrases and entities to learn knowledge information in these masking units.
As a reward, syntactic and semantic information from phrases and entities is implicitly integrated into the language model.
SpanBERT~\cite{joshi2020spanbert} extends subword-level masking to span-level masking, and selects random spans of full words to predict.
In addition to the MLM objective, a span boundary objective is designed to predict each subword within a span using subword representations at the boundaries.
\citet{joshi2020contextualized} utilize retrieved background sentences for phrases to extend the input text, and combine the TEK-enriched context and span masking to pre-train the language model.
Besides, \citet{rosset2020knowledge} introduce entity information into an autoregressive language model, called KALM, which identifies the entity surface in a text sequence and maps word n-grams into entities to obtain an entity sequence for knowledge-aware language model pre-training. 

\citet{ye2019align} propose a discriminative pre-training approach for incorporating commonsense knowledge into the language model, in which the question is concatenated with different candidates to construct a multi-choice question answering sample, and each choice is used to predict whether the candidate is the correct answer.
KEPLER~\cite{wang2019kepler} unifies knowledge representation learning and language modeling objectives, which builds a bridge between text representation and knowledge embeddings by encoding entity descriptions, and can better integrate factual knowledge into the pre-trained language model.
\citet{xiong2019pretrained} introduce entity replacement checking task into the pre-trained language model, which greatly enhances the modeling of entity information.
\citet{wang2020k} propose a plug-in way to infuse knowledge into language models, and different discriminative objectives are used to keep different kinds of knowledge in different adapters.

\begin{figure*}[t]
	\centering
	\includegraphics[trim={0 0in 0in 5in}, clip, width=\linewidth]{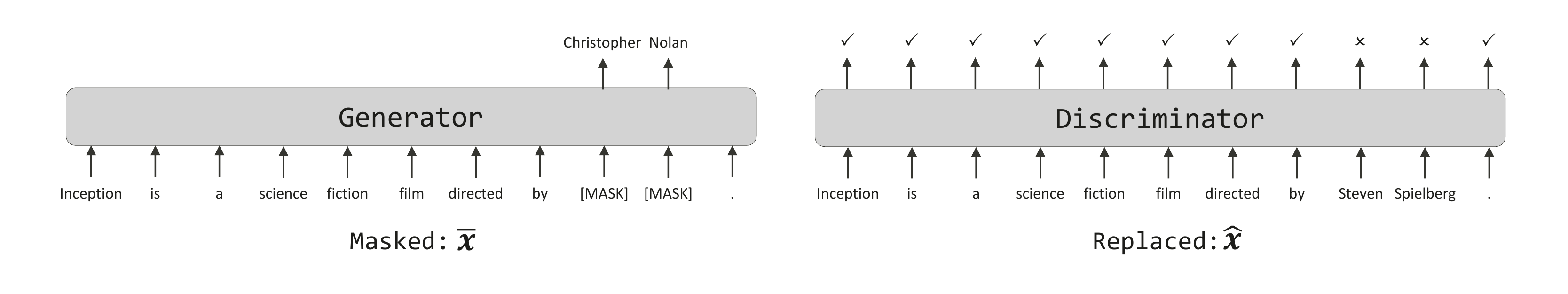}
	\caption{\label{fig:model} Knowledge-guided pre-training with generative and discriminative objectives. The knowledge span \textit{Christopher Nolan} is masked for token prediction in the Generator (left). A ``fake'' span \textit{Steven Spielberg} replacing the original span \textit{Christopher Nolan} is fed to the Discriminator (right) to predict whether the tokens are replaced or not.}
\end{figure*}

To the best of our knowledge, this work is the first to explore knowledge-guided pre-training that considers the \textit{generative} and \textit{discriminative} approaches simultaneously. Besides, our model does not involve any additional cost to downstream tasks in terms of parameters and computations.


\section{Methodology}

\subsection{Knowledge-guided Pre-training}

The vanilla masked language model performs pre-training by maximizing the log-likelihood:
\begin{equation}\label{mlm}
\begin{split}
\mathcal{L}_{\mathrm{MLM}} (\boldsymbol{\theta}) = \sum_{i = 1}^n m_i \log p(x_i|\boldsymbol{x}\ccirc \boldsymbol{m};\boldsymbol{\theta}),
\end{split}
\end{equation}
where $\boldsymbol{x}=[x_1,\ldots,x_n]$ is the original sequence of tokens, $\ccirc$ denotes the element-wise product, and $\boldsymbol{m}=[m_1,\ldots,m_n]$ is the binary vector where $m_i=1$ indicates replacing $x_i$ with \texttt{[MASK]} or otherwise unchanged. $\boldsymbol{\theta}$ represents the model parameter.

For knowledge-guided pre-trained language model, there are two approaches to incorporate information from knowledge sources.
One is the \textit{generative} approach that pushes model to predict the masked knowledge-sensitive tokens, such as tokens in entities or relations. Formally, with the masking scheme $\boldsymbol{m}^{\mathcal{K}}$ induced by knowledge sources $\mathcal{K}$, the masked sequence becomes
$\bar{\boldsymbol{x}}=\texttt{K-MASK}(\boldsymbol{x})=\boldsymbol{x}\ccirc \boldsymbol{m}^{\mathcal{K}}$.
The generative loss is then defined as the negative log-likelihood of predicting the masked tokens:
\begin{equation}\label{gen}
\begin{split}
\mathcal{L}_{\mathrm{Gen}}(\boldsymbol{\theta}_g) = -\sum_{i = 1}^n m^{\mathcal{K}}_i \log p(x_i|\bar{\boldsymbol{x}};\boldsymbol{\theta}_g).
\end{split}
\end{equation}

The other is the \textit{discriminative} approach where the original sequence is converted to $\hat{\boldsymbol{x}}=\texttt{K-Replace}(\boldsymbol{x})$ by replacing the knowledge-sensitive parts of sequence with other related texts according to the knowledge contained in $\mathcal{K}$. Discriminative loss could be defined at sequence-level, span-level, or token-level. In this paper, we use the fine-grained token-level loss, which is the cross-entropy loss of classifying whether the token is replaced or not:  
\begin{equation}\label{disc}
\begin{split}
\mathcal{L}_{\mathrm{Disc}}(\boldsymbol{\theta}_d) = - &\sum_{i=1}^n \mathbbm{1}(\hat{x}_i=x_i) \log p(r_i=0|\hat{\boldsymbol{x}};\boldsymbol{\theta}_d)\\
 & + \mathbbm{1}(\hat{x}_i\neq x_i) \log p(r_i=1|\hat{\boldsymbol{x}};\boldsymbol{\theta}_d).
\end{split}
\end{equation}
Here each $r_i$ is a binary variable indicating whether $\hat{x}_i$ is different from $x_i$.

Figure~\ref{fig:model} illustrate the diagram of the generator and the discriminator from the two perspectives. The proposed knowledge-guided pre-training is defined as the following optimization problem:

\begin{equation}\label{kglm}
\begin{split}
\min_{\boldsymbol{\theta}_g, \boldsymbol{\theta_d}}  \mathcal{L}_{\mathrm{Gen}}(\boldsymbol{\theta}_g) + \lambda \mathcal{L}_{\mathrm{Disc}}(\boldsymbol{\theta}_d).
\end{split}
\end{equation}

In the following sections, we detail our approach of using external knowledge to design the generative (section~\ref{sec:gen}) and discriminative tasks (section~\ref{sec:disc}) respectively, and explore different ways of joint learning of the two models (section~\ref{sec:param}).

\subsection{Generative Pre-training}\label{sec:gen}

Given the input sequence $\boldsymbol{x}$, unlike the conventional random masking, we particularly extract spans of text which are more knowledge intensive as the candidates for masking. In this work, we use Wikipedia text as the pre-training corpus, and follow the approach proposed in WiNER~\cite{ghaddar2017winer} to extract the knowledge-sensitive spans. Briefly, the anchored strings in a Wikipedia page are first collected. Then, through mining titles of out-linked pages and their co-referring mentions in the current page, more salient spans are identified. As reported in~\cite{ghaddar2017winer}, the coverage of the knowledge-sensitive spans (mostly the entities and concepts) in the Wikipedia texts can be increased to 15\%.

As the example shown in the left part of Figure~\ref{fig:model}, entities or concepts such as ``\textit{Inception}'', ``\textit{science fiction film}'' and ``\textit{Christopher Nolan}'' can be extracted from the sentence. In generative pre-training, candidate spans are randomly selected and masked (``\textit{Christopher Nolan}'' in the example) to produce the masked sequence $\bar{\boldsymbol{x}}$. Then the generator is trained to recover the masked tokens. Here we use a 12-layer Transformer encoder as the architecture for the generator.

In practice, since the coverage of the knowledge span can vary in different sentences, we use a mixture of span masking and subword masking, while still keeping the total masking ratio of 15\%. Specifically, a probabilistic masking strategy is designed to choose whether to mask a knowledge span or a subword every time. For each sentence, knowledge spans (if available) or subwords are masked iteratively by following the strategy until reaching the masking ratio of 15\%. We denote the whole knowledge masking procedure as \texttt{K-MASK} in the paper.

\subsection{Discriminative Pre-training}\label{sec:disc}

We use the same knowledge spans extraction method as in generative pre-training. Once the spans are identified, we randomly replace them with other pieces of text, and fed the altered sequences $\hat{\boldsymbol{x}}$ to the discriminator to distinguish which tokens are replaced ($r_i=1$) and which are not ($r_i=0$).

For example, in the right part of Figure~\ref{fig:model}, the span ``\textit{Christopher Nolan}'' in the sentence is selected and replaced by ``\textit{Steven Spielberg}'', and the model learns to detect the ``fake'' tokens in the sequence.

For each token, the probability of being replaced is computed as
\begin{equation}
\label{equ:disc}
p(r_i|\hat{\boldsymbol{x}};\boldsymbol{\theta}_{d})=\mathrm{sigmoid}(w^T\mathrm{GeLU}(\boldsymbol{W}\boldsymbol{h}(\hat{x}_i)),
\end{equation}
where $\boldsymbol{h}(\hat{x}_i)$ is the contextualized representation of $\hat{x}_i$ by the discriminator, which is also a 12-layer Transformer encoder.

In order to make the model more discriminative to the factual knowledge, we replace the knowledge spans with other spans of the same type. In addition to the knowledge span substitution, we also randomly replace some of the subwords with random tokens. The total replacement ratio of a sentence is kept at 15\%. We denote the whole knowledge span replacement process as \texttt{K-Replace} in the paper.

\begin{figure}[!t]
	\centering
	\begin{subfigure}{0.49\linewidth}
	\centering
	\includegraphics[trim={9.5in 0.1in 0in 2.9in}, clip, width=\linewidth]{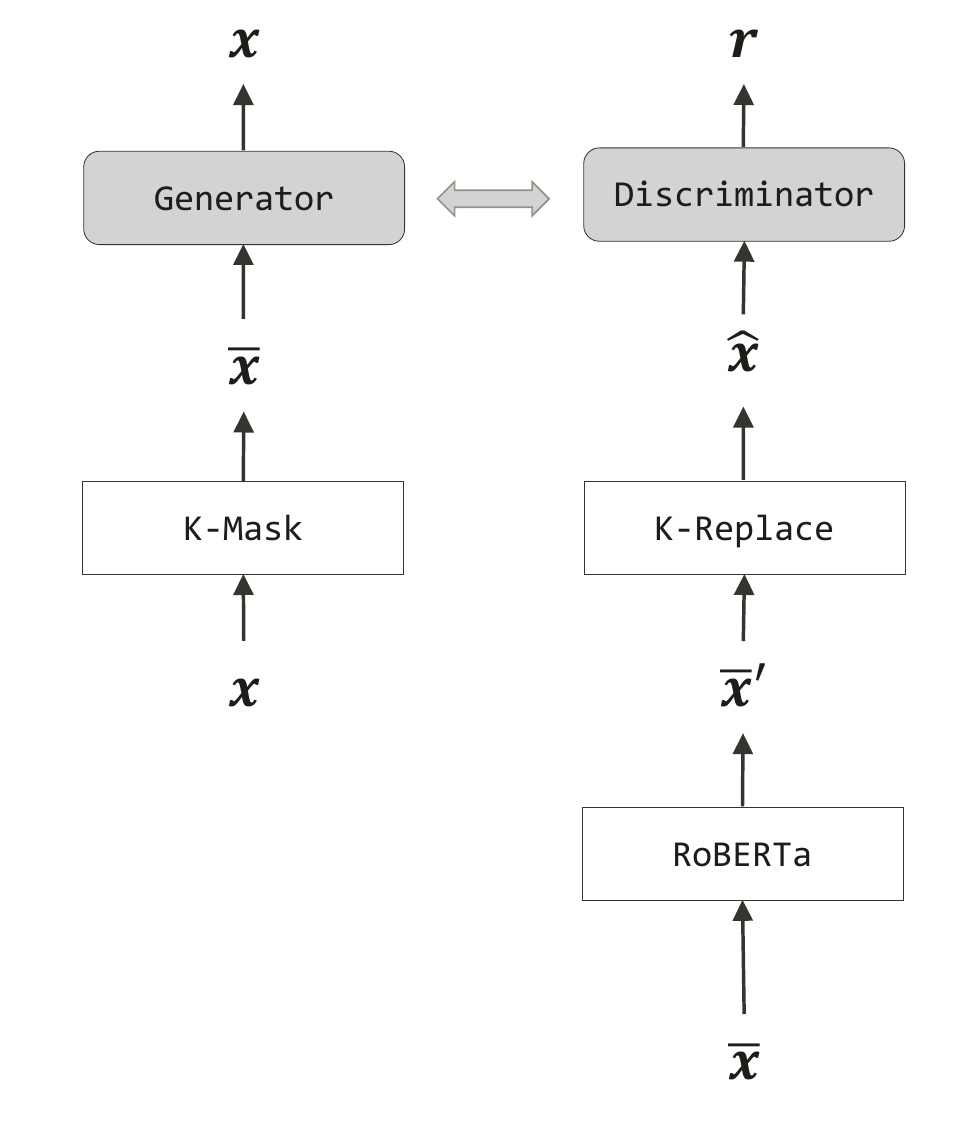}
	\caption{Two-tower}\label{fig:two-tower}
	\end{subfigure}
	\begin{subfigure}{0.49\linewidth}
	\centering
	\includegraphics[trim={9.5in 0in 0in 3in}, clip, width=\linewidth]{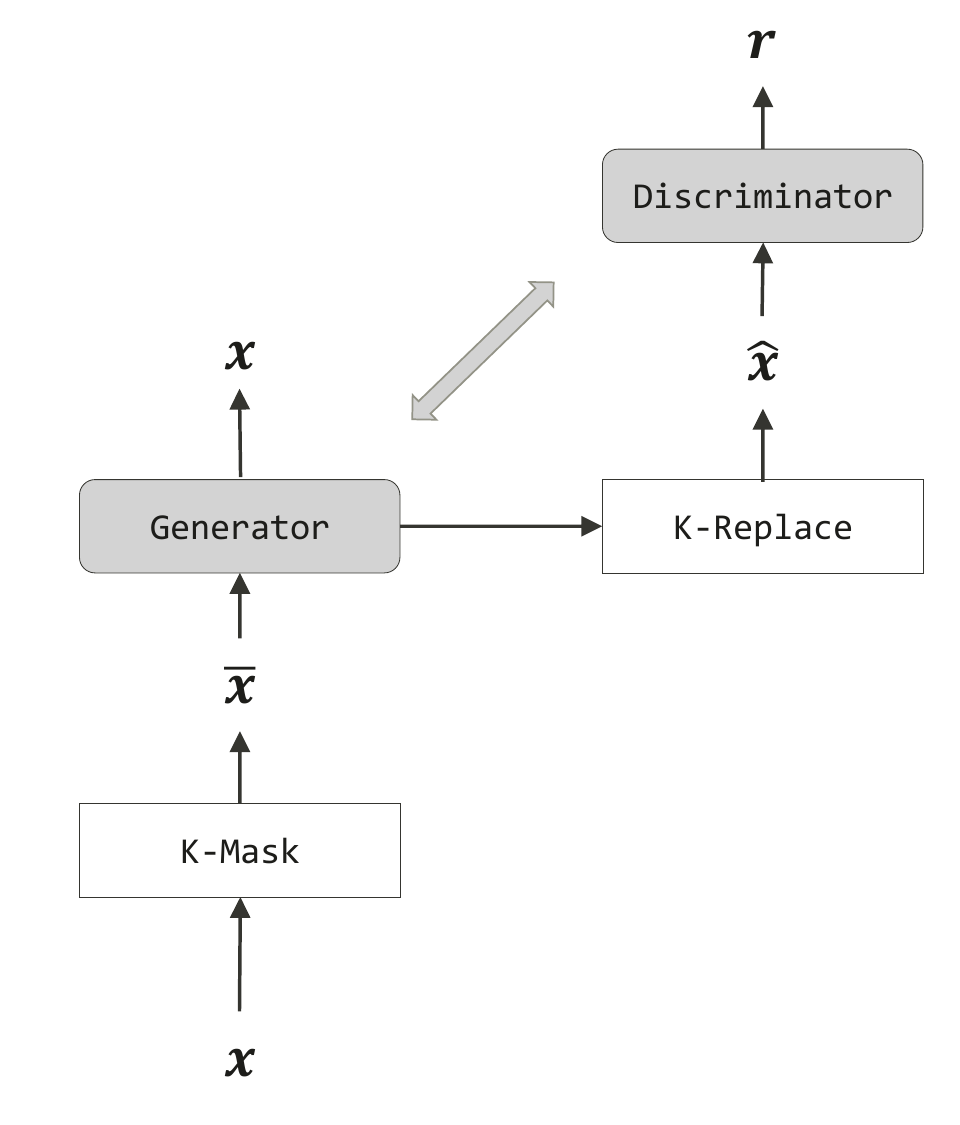}
	\caption{Pipeline}\label{fig:pipeline}
	\end{subfigure}
	\caption{\label{fig:schemes} Knowledge-guided pre-training schemes. \texttt{K-MASK} refers to the knowledge span masking and \texttt{K-Replace} refers to the knowledge span replacement. Generator works for masked words prediction and discriminator works for span replacement detection. Double-sided arrow between the generator and discriminator means their parameters can be shared.}
\end{figure}

\subsection{Learning Schemes}\label{sec:param}

In this work, generative and discriminative pre-training are combined to promote the integration of knowledge into the language model. We design two learning schemes, which are referred to as the \textit{two-tower} scheme and the \textit{pipeline} scheme, respectively.

Figure~\ref{fig:schemes}(a) illustrates the two-tower architecture, where the generator and discriminator are simply trained in parallel, and their Transformer parameters are shared. 

Figure~\ref{fig:schemes}(b) shows the pipeline model, which connects the output of the generator to the discriminator. More precisely, the original sequence is first masked and then recovered by the generator. After applying knowledge span replacement on the recovered sequence, the discriminator learns to tell which tokens are changed from the original input, including those by the generator and the span replacement. The parameter can also be shared across generator and discriminator.

In this paper, we denote KgPLM$_{T}$ for the two-tower model and KgPLM$_{P}$ for the pipeline model.

\begin{table}
	\centering
	\small
	\begin{tabularx}{\linewidth}{lrCC}
		\hline
		\bf Dataset & \bf Facts & \bf Relations & \bf Subwords \\
		\hline
		LAMA-Google-RE & 4618 & 3 & $=$1\\
		LAMA-T-REx & 29521 & 41   & $=$1\\
		LAMA-UHN-Google-RE & 4102 & 3 & $=$1\\
		LAMA-UHN-T-REx & 23220 & 41   & $=$1\\
		\hline
		LAMA-Google-RE$^{*}$ & 909 & 3 & $>$1\\
		LAMA-T-REx$^{*}$  & 4514 & 41  & $>$1\\
		LAMA-UHN-Google-RE$^{*}$ & 829 & 3 & $>$1\\
		LAMA-UHN-T-REx$^{*}$  & 3886 & 40  & $>$1\\
		\hline
	\end{tabularx}
    \caption{Datasets for cloze-style question answering. 
		``Subwords'' means the number of subwords within the answer.
		$*$ denotes that samples with more than one subword in the answer.}\label{tab:cqa_data}
\end{table}

\section{Experiments}

\subsection{Pre-training Setup}

Pre-training large-scale language model from scratch is expensive, so we use the  RoBERTa$_{\text{BASE}}$ implemented by Huggingface\footnote{\url{https://github.com/huggingface/transformers}.} as the initialization of our model.
We follow the model configuration in RoBERTa$_{\text{BASE}}$ and continue the knowledge-guided pre-training on English Wikipedia\footnote{\url{https://hotpotqa.github.io/wiki-readme.html}.} for 5 epochs. 
Details of the knowledge span extraction from Wikipedia is shown in Sec~\ref{subsec:span}.
Different from randomly masking a certain number of subwords in general pre-training, knowledge-related spans are not uniformly distributed in the corpus. Therefore, we use \textit{dynamic span masking} to select the knowledge spans, thus different spans are learned in different pre-training epochs.
Furthermore, we employ \textit{dynamic span replacement} to choose different negative knowledge spans for each epoch.
The details of pre-training experimental settings are described in Sec~\ref{subsec:pretrain}.

\subsection{Cloze-style Question Answering}
\label{subsec:cqa}

\begin{table*}
	\centering
	\small
	\begin{tabularx}{\linewidth}{lccccCCC}
		\hline
		&  \multicolumn{3}{c}{\textbf{LARGE Models}} & \multicolumn{4}{c}{\textbf{BASE Models}}  \\ \cmidrule(lr){2-4} \cmidrule(lr){5-8}
		\textbf{Dataset} & RoBERTa$_{\text{LARGE}}$$^{\dag}$ & K-ADAPTER$^{\dag}$ & RoBERTa$_{\text{LARGE}}$ & RoBERTa$_{\text{BASE}}$ & WKLM$^\ddag$ & KgPLM$_T$ & KgPLM$_P$  \\
		\hline
		LAMA-Google-RE & 4.8 & 7.0 & 4.8 & 5.3 & 6.0 & \bf 10.5 & 9.2 \\
		LAMA-T-REx & 27.1 & 29.1  & 28.1 & 21.7 & 22.5 & \bf 31.1 & 27.9 \\
		LAMA-UHN-Google-RE & 2.5 & 3.7 & 2.5 & 2.2 & 3.0 & \bf 5.0 & 4.9\\
		LAMA-UHN-T-REx & 20.1 & 23.0 & 21.1 & 14.5 & 16.6 & \bf 22.7 & 20.4 \\
		\hline
		LAMA-Google-RE$^{*}$ & - & - & 0.4 & 0.1 & 0.0 & \bf 6.1 & 5.7 \\
		LAMA-T-REx$^{*}$  & - & -  & 8.8 & 5.2 &1.3 & \bf 18.5 & 15.8 \\
		LAMA-UHN-Google-RE$^{*}$ & - & - & 0.4 & 0.1 & 0.0 & \bf 2.7 & \bf 2.7 \\
		LAMA-UHN-T-REx$^{*}$  & - & -  &  2.1 & 0.7 & 0.2 & \bf 9.0 & 7.5 \\
		\hline
	\end{tabularx}
    \caption{Performance (Macro P@1) on cloze-style question answering of factual knowledge. $*$ denotes that samples with only one subword in the answer are filtered. ${\dag}$ denotes that these results are copied from the original paper~\cite{wang2020k}. $^\ddag$ means the model is re-implemented and trained by us.}\label{tab:cqa}
\end{table*}

\subsubsection{Datasets}

To examine the model's ability to memorize and complete the factual knowledge stored in natural language statements, we use cloze-style question answering benchmark LAMA\footnote{\url{https://github.com/facebookresearch/LAMA}.}~\cite{petroni2019language} and LAMA-UHN~\cite{poerner2019bert} to evaluate our models. %
LAMA-UHN is a subset of LAMA, which is constructed by filtering out samples in LAMA that are easy to answer from entity names alone.
On these datasets, we can simply use masked subwords prediction to answer questions.
For example, the question is ``\textit{Christopher Nolan was born in [MASK] .}'', and the target is to predict the masked answer ``\textit{London}''.
We do our experiments on LAMA-Google-RE, LAMA-T-REx, LAMA-UHN-Google-RE and LAMA-UHN-T-REx, which focus on factual knowledge.
In our models, a byte-level Byte-Pair Encoding (BPE) vocabulary is used to tokenize the input sequence, which makes some answers are split into multiple subwords.
In order to evaluate the models under the multi-subword setting, we filter samples with only one subword in the answer and construct LAMA-Google-RE$^{*}$, LAMA-T-REx$^{*}$, LAMA-UHN-Google-RE$^{*}$ and LAMA-UHN-T-REx$^{*}$.
The statistics of these datasets are listed in Table~\ref{tab:cqa_data}.
Similar to \cite{wang2020k}, we remove stop words from the ranking vocabulary.

\subsubsection{Settings}
We use precision at one (P@1) to evaluate the predictions. For the multi-subword answer, the sample is considered as correct when all the subwords within the answer are predicted correctly.
Macro-averaged P@1 over different relations is reported as the final score.
In our experiments, data preprocessing and hyperparameter settings remain the same as~\cite{petroni2019language}. 

\subsubsection{Baselines}

We compare our models with RoBERTa and two knowledge-guided pre-trained language models:

\begin{itemize}
\item \textbf{RoBERTa}~\cite{liu2019roberta} Compared with BERT, RoBERTa replaces the static masking strategy with \textit{dynamic masking}, and is pre-trained over more data and longer steps. Moreover,  the \textit{next sentence prediction} pre-training task is removed.

\item \textbf{K-ADAPTER}~\cite{wang2020k} Knowledge triples are aligned with natural text through a distant supervision method, and these natural text is utilized to inject knowledge into the language model by a relation classification task.

\item \textbf{WKLM}~\cite{xiong2019pretrained} We re-implement the WKLM model, which supplements the pre-training objective by identifying whether an entity in a certain context is replaced or not. Unlike that in the original paper, we use the more powerful model RoBERTa$_{\text{BASE}}$ to be the initialization.
We follow the pre-training settings used in~\cite{xiong2019pretrained}, and continue the corresponding knowledge-guided pre-training on English Wikipedia for 20 epochs.
\end{itemize}

\begin{table}[!t]
	\centering
	\begin{tabular}{lrr}
		\hline
		\textbf{Datasets} & \textbf{\# Train} & \textbf{\# Dev} \\
		\hline
		SQuAD &  86,588  &  10,507  \\
		NewsQA &  74,160 & 4,212 \\
		TriviaQA &  61,688 &  7,785  \\
		SearchQA &  117,384  & 16,980  \\
		HotpotQA  & 72,928  & 5,901  \\
		Natural Questions  & 104,071  & 12,836  \\
		\hline
		Total & 516,819 & 58,221 \\
		\hline
	\end{tabular}
	\caption{MRQA datasets.}\label{tab:mrqa_data}
\end{table}

\begin{table*}
	\centering
	\begin{tabularx}{\linewidth}{lCCCCCCCCCC|CC}
		\hline
		&  \multicolumn{2}{c}{\textbf{NewsQA}} & \multicolumn{2}{c}{\textbf{TriviaQA}} & \multicolumn{2}{c}{\textbf{SearchQA}}  & \multicolumn{2}{c}{\textbf{HotpotQA}} & \multicolumn{2}{c|}{\textbf{NaturalQA}} & \multicolumn{2}{c}{\textbf{Avg.}} \\
		\textbf{Model} & EM & F1 & EM & F1 & EM & F1 & EM & F1 & EM & F1& EM & F1  \\
		\hline
		BERT$_{\text{LARGE}}$$^{\dag}$ & - & 68.8 & - & 77.5 & - & 81.7 & - & 78.3 & - & 79.9 & - & 77.3 \\
		SpanBERT$_{\text{LARGE}}$$^{\dag}$ & - & 73.6 & - & 83.6 & - & 84.8 & - & 83.0 & - & 82.5 & - & 81.5 \\
		\hline
		SpanBERT$_{\text{BASE}}$  & 56.27 & 70.92 & 74.93 & 79.77 & 77.99 & 83.73 & 64.78 & \bf 80.10 & 69.07 & 80.35 & 68.61 & 78.97 \\
		RoBERTa$_{\text{BASE}}$  & 57.06 & 71.11 & 73.80 & 78.18 & 78.56 & 84.11 & 63.97 & 79.37 & 68.46 & 79.97 & 68.37 & 78.55 \\
	    WKLM$^\ddag$ & 57.16 & 71.51 & 74.93 & 79.45 & 78.73 & 84.25 & 63.83 & 79.60 & 69.35 & 80.54 & 68.80 & 79.07 \\
		KgPLM$_{T}$ (5 epochs) & 57.11 & 71.69 & 74.93 & 79.67 & 79.07 & 84.55 & 64.61 & 79.76 & 69.88 & \bf 80.99 & 69.12 & 79.33 \\
		KgPLM$_{P}$ (5 epochs) &  57.30 & \bf 72.37 &  75.37 & 79.74 & 79.02 & \bf 84.60 & 64.20 & 80.03 & 69.87 & 80.95 &  69.15 & \bf 79.54 \\
		KgPLM$_{P}$ (10 epochs)  & 57.65 & 71.86 & 75.42 & \bf 79.93 & 78.82 & 84.35 & 64.75 & 79.97 & 69.79 & 80.94 & 69.29 & 79.41 \\
		\hline
	\end{tabularx}
	\caption{Test results of the \textit{MRQA-single} setting. ${\dag}$ denotes that these results are copied from the original paper~\cite{joshi2020spanbert}. $^\ddag$ means the model is re-implemented and trained by us. ``NaturalQA'' is short for ``Natural Questions'', hereinafter referred to as ``NaturalQA''.}\label{tab:mrqa}
\end{table*}

\subsubsection{Results}

Table~\ref{tab:cqa} shows the results on cloze-style question answering of factual knowledge.
We observe that KgPLM outperforms RoBERTa$_{\text{BASE}}$ by a large margin, even outperforms the LARGE models.
For LAMA-Google-RE and LAMA-T-REx, WKLM improves 0.7 and 0.8 on P@1 over RoBERTa$_{\text{BASE}}$, while KgPLM$_T$ achieves much larger gains of 5.2 and 9.4 on P@1.
Moreover, compared to RoBERTa$_{\text{BASE}}$ and WKLM, KgPLM performs much better on LAMA-UHN datasets.
This indicates that KgPLM really captures richer factual knowledge than the baseline models.
We also check the performance comparison in each relation type (not listed due to the limited space) and find that KgPLM outperforms RoBERTa in most relation types, indicating that KgPLM improves over a wide range of factual knowledge.

All the answers in LAMA$^{*}$ and LAMA-UHN$^{*}$ contain more than one subword, and these datasets can be used to detect model's ability on capturing span-level knowledge.
From Table~\ref{tab:cqa}, we can see that the P@1 scores of RoBERTa on LAMA$^{*}$ and LAMA-UHN$^{*}$ drop dramatically, comparing to those on LAMA and LAMA-UHN.
WKLM also works not well on LAMA$^{*}$ and LAMA-UHN$^{*}$, and one of the possible reasons is that the span-level pre-training objective in WKLM is discriminative but cloze-style question answering is a generative task.
Similarly, the performance of KgPLM decreases but with much smaller decline, and eventually KgPLM$_T$ exceeds 6.0, 13.3, 2.6 and 8.3 P@1 score over RoBERTa on LAMA-Google-RE$^{*}$, LAMA-T-REx$^{*}$, LAMA-UHN-Google-RE$^{*}$ and LAMA-UHN-T-REx$^{*}$, respectively.
These experimental results demonstrate that span-level knowledge-guided pre-training indeed enhances the model from perspective of span-level knowledge completion.

\subsection{Machine Reading for Question Answering}
\label{subsec:mrqa}

\subsubsection{Datasets} 
We evaluate the knowledge-guided language models on six question answering datasets from the MRQA shared task~\cite{fisch2019mrqa}.
These datasets follow the task setting of machine reading comprehension: given a question and a paragraph, the objective is to extract a text span from the paragraph to be the final answer.
In this paper, we train our models on the in-domain training sets and evaluate on the in-domain development sets.

\subsubsection{Settings} 
Table~\ref{tab:mrqa_data} lists the data statistics, and the experiments are conducted in two settings:
(1) \textit{MRQA-single}, we follow the dataset setting used in~\cite{joshi2020spanbert}, which splits the in-domain development set in half to generate the test set, and individual models are trained and evaluated on each dataset;
(2) \textit{MRQA-combine}, the training sets of different datasets are combined to train a single model, and the best checkpoint is selected by evaluating the entire development set.
Exact match (EM) and word-level F1 scores are used to evaluate model performance.
For the first finetuning setting, by referring to~\cite{joshi2020spanbert}, we search the best hyperparameters  for each dataset.
For the second setting, because the training set is large,  the chosen hyperparameters in~\cite{joshi2020contextualized} are directly utilized to finetune our models.
See Sec~\ref{subsec:finetune} for finetuning details.

\subsubsection{Baselines}
In addition to RoBERTa and WKLM (mentioned in \ref{subsec:cqa}), we compare our models with two other baselines, and all these models are \texttt{BASE} models:
\begin{itemize}
	\item \textbf{SpanBERT}~\cite{joshi2020spanbert} This model extends subword-level masking to span-level masking, and selects random spans of full words to predict. In addition to the MLM objective, a span boundary objective is designed to predict each subword within a span using subword representations at the boundaries. In this work, the released \texttt{spanbert-base-cased} model\footnote{\url{https://github.com/facebookresearch/SpanBERT/}.} is utilized to do the experiments. %
	\item \textbf{TEK}~\cite{joshi2020contextualized} Retrieved background sentences for phrases in the context, called ``textual encyclopedic knowledge'', are utilized to extend the input text. The TEK-enriched context can be used for model pre-training and finetuning.
\end{itemize}

\begin{table}
	\centering
	\begin{tabular}{lcc}
		\hline
		\textbf{Model} & \bf EM & \bf F1 \\
		\hline
		RoBERTa$_{\text{BASE}}$$^{\dag}$ &  - & 82.98 \\
		RoBERTa$_{\text{BASE}}$++$^{\dag}$ &  - & 83.2 \\
		TEK$_{F}$$^{\dag}$ &  - & 83.44 \\
		TEK$_{P}$$^{\dag}$ &  - & 83.3 \\
		TEK$_{PF}$$^{\dag}$ &  - & 83.71 \\
		\hline
		KgPLM$_{T}$ (5 epochs) & 75.32 & \bf 83.72 \\ 
		KgPLM$_{P}$ (5 epochs) & 75.29 & \bf 83.72 \\ 
		KgPLM$_{P}$ (10 epochs) & 75.20 & 83.70 \\ 
		\hline
	\end{tabular}
	\caption{Dev results of the \textit{MRQA-combine} setting. ${\dag}$ denotes that these results are copied from the original paper~\cite{joshi2020contextualized}.}\label{tab:mrqa_dev}
\end{table}

\subsubsection{Results}

\begin{table*}[t]
	\centering
	\begin{tabular}{lccccc|c}
		\hline
		\bf Model &\textbf{NewsQA} & \textbf{TriviaQA} & \textbf{SearchQA}  & \textbf{HotpotQA} & \textbf{NaturalQA} & \textbf{Avg.} \\
		\hline
		KgPLM$_T$  & 71.69  & 79.67  & 84.55  & 79.76 & \bf 80.99 & 79.33 \\ 
		\ \ \ \ w/o \texttt{K-Replace}  & 71.88 & 79.53 & 84.35  & 79.68 & 80.73  & 79.23 \\
		\hline
		KgPLM$_P$  & \bf 72.37  & \bf 79.74  & \bf 84.60  & \bf 80.03  & 80.95  & \bf 79.54 \\
		\ \ \ \ w/o \texttt{K-Replace}  & 71.56 & 79.33  & 84.18 & 79.51 & 80.94  & 79.10 \\
		\hline
	\end{tabular}
    \caption{Performance (F1) on MRQA test sets for different pre-training schemes.}\label{tab:schemes}
\end{table*}

\begin{table*}
	\centering
	\begin{tabular}{lccccc|c}
		\hline
		\textbf{Model} & \textbf{NewsQA} & \textbf{TriviaQA} & \textbf{SearchQA}  & \textbf{HotpotQA} & \textbf{NaturalQA} & \textbf{Avg.} \\
		\hline
		RoBERTa$_{\text{BASE}}$  & 71.11  & 78.18  & 84.11  & 79.37  & 79.97  & 78.55 \\
		RoBERTa$_{\text{BASE}}^+$ & 71.43 & 79.31  & 84.27 & 79.89 & 80.51 & 79.08 \\
		\hline
		KgPLM$_{P}$   & \bf 72.37 & \bf 79.74  & \bf 84.60& \bf 80.03  & \bf 80.95 & \bf 79.54 \\
        \ \ \ \ w/o $\mathcal{L}_{\mathrm{Gen}}$ & 71.92 & 78.40  & 84.24 & 79.81  & 80.70 & 79.01 \\
		\ \ \ \ w/o $\mathcal{L}_{\mathrm{Disc}}$ & 71.99 & 78.74 & 84.11  & 79.71 &  80.91  & 79.09 \\
		\hline
	\end{tabular}
    \caption{Ablation study of continue general MLM pre-training and do knowledge-guided pre-training with different objectives.}\label{tab:continue}
\end{table*}

Table~\ref{tab:mrqa} shows the test results of the \textit{MRQA-single} setting.
RoBERTa falls behind the previous state-of-the-art model SpanBERT.
In these five datasets, TriviaQA shows the most significant performance gap, where SpanBERT exceeds RoBERTa 1.59 F1 score.
92.85\% of the answers in TriviaQA are Wikipedia titles~\cite{joshi2017triviaqa} and the pre-training steps of SpanBERT on Wikipedia are much larger than RoBERTa, which make SpanBERT so well-performed.
Based on a knowledge-guided discriminative objective, our reproduced WKLM improves over the RoBERTa baseline, achieving a competitive performance with SpanBERT.
Among the \texttt{BASE} models, KgPLM outperforms all the baselines in averaged F1 score.
Compared with RoBERTa, KgPLM$_{P}$ achieves improvements in each dataset, increasing the F1 score by 1.26, 1.56 and 0.98 on NewsQA, TriviaQA and Natural Questions, respectively.
These experimental results indicate that our knowledge-guided pre-training method (combination of \textit{knowledge span masking} and \textit{knowledge span replacement checking}) enhances the model's capacity in absorbing knowledge and benefits the downstream QA tasks.

Table~\ref{tab:mrqa_dev} shows the dev results of the \textit{MRQA-combine} setting.
Comparing with TEK models, KgPLM achieves comparable F1 score on the combined development set of MRQA datasets.
In \cite{joshi2020contextualized}, the TEK-enriched context benefits pre-training and finetuning.
In our work, we haven't introduce any data augmentation in pre-training, so the the TEK-enriched context will presumably also help our models. If we enrich the inputs with TEK in our pre-training and finetuning procedure, there may still exists some room for improvement based on the current model. We leave this for our future work.

\section{Discussion}

\subsection{Pre-training Schemes}

In addition to the two-tower and pipeline pre-training schemes proposed in Sec~\ref{sec:param}, we explore another two schemes by removing the \texttt{K-Replace} module.
Without knowledge span replacement, the output of the generator in the pipeline-based model is fed into the discriminator directly, which brings a great challenge to the learning of the discriminator.
The intermediate sequence recovered by the generator may destroy the grammatical structure of the sentence while corrupting the integrity of the knowledge span itself, which is not a good learning sample for the discriminator.
As shown in Table~\ref{tab:schemes}, this scheme (Row 4) achieves much lower averaged score on MRQA test sets than the original pipeline scheme (Row 3).
The comparison of the two-tower-based models (Row 1 and 2) also shows that the \texttt{K-Replace} module is beneficial to the learning of knowledge.
These experimental results demonstrate the importance of the \texttt{K-Replace} module in our models.
Knowledge span replacement ensures the integrity of negative knowledge spans in the input, and do not destroy the grammatical structure of the input sentence.
Moreover, the uncorrupted negative knowledge spans improve the learning effectiveness of \textit{knowledge span replacement checking} in the discriminator.

\subsection{Continue Pre-training}

To make a fair comparison with knowledge-guided pre-training, we continue pre-training based on the released \texttt{roberta-base} checkpoint via the general MLM objective on Wikipedia.
We call this extended model RoBERTa$_{\text{BASE}}^+$.
Table~\ref{tab:continue} shows comparison between do knowledge-guided pre-training and continue general pre-training.
From row 1 and 2, we can see that continue pre-training on Wikipedia improves averaged F1 score of the MRQA test sets by 0.53, which gains largest on TriviaQA, followed by NaturalQA and HotpotQA.
Compared with RoBERTa$_{\text{BASE}}^+$, KgPLM exhibits better performance on all datasets, indicating that our method is more effective than general MLM in learning knowledge from text. 

\subsection{Effects of Generator and Discriminator}

We conduct further analysis to differentiate the effects of the generator and the discriminator in our models.
Based on the pipeline-based model KgPLM$_P$, we train another two models: KgPLM$_P$ w/o the generator, and KgPLM$_P$ w/o the discriminator.
The ablation results are shown in Table~\ref{tab:continue}.
After removing each knowledge injection module, the model performance decreases, which indicates that these two knowledge-guided pre-training tasks are complementary to each other in knowledge consolidation.

\section{Conclusion}

We have proposed a pre-training method by cooperatively modeling the generative and discriminative knowledge injecting approaches.
Our model can be easily extended to larger pre-training corpus and does not introduce any modifications for downstream tasks during finetuning.
Experiments show our model consistently outperforms all \texttt{BASE} models on a variety of question answering datasets, demonstrating that our KgPLM is a preferred choice for the knowledge intensive NLP tasks.


%

%

\bibliography{kgplm}
\bibliographystyle{aaai21}

\clearpage
\appendix

\section{Appendices}\label{sec:appendix}

\subsection{Knowledge Span}\label{subsec:span}

Inspired by WiNER \cite{ghaddar2017winer}, we extract knowledge spans for each article in Wikipedia by the following steps:
\begin{enumerate}[1.]
	\item Given an article $D$, all the anchored strings are extracted as knowledge spans, then all these anchored strings are added into a span set $S$.
	\item All these anchors have an out-linked article. We collect the anchored strings in these out-linked articles (2-hop anchor) and add them into $S$.
	\item Most Wikipedia title has a Wikidata item, and aliases may exist for this item. We extend aliases for each knowledge span in $S$, and utilize the final span set to extract knowledge spans in the article $D$.
\end{enumerate}
After extracting knowledge spans for each article in Wikipedia, the statistics are shown in Table~\ref{tab:ks}.

\begin{table}[!h]
	\centering
		\small
	\begin{tabular}{lccc}
		\hline
		\bf Span Source & \bf  \# Article & \bf \# KS & \bf Avg./Max. \\
		\hline
		Anchor & 5,486,211 & 97,167,509 & 17.7/9069 \\
		\ \ + 2-hop  & 5,486,211 & 133,886,575 & 24.4/9379 \\
		\ \ + Alias & 5,486,211 & 146,865,593 & 26.8/9379 \\
		\hline
	\end{tabular}
	\caption{Statistics of knowledge spans in Wikipedia. KS, knowledge span.}\label{tab:ks}
\end{table}

\subsection{Pre-training Details}
\label{subsec:pretrain}

As in RoBERTa, we set the max sequence length to 512, and continue pre-training for 5 epochs on English Wikipedia.
The weight for the discriminative objective $\lambda$ is set to 25.
We use the AdamW~\cite{kingma2014adam} optimizer with the learning rate of 2e-5.
The sampling ratio to choose whether to mask a knowledge span or a subword is set to 0.5.
To accelerate the pre-training process, we follow the \texttt{balanced data parallel} strategy used in Transfomer-XL~\cite{dai2019transformer} and pre-train the models on 8 V100 GPUs with a batch size of 84.
The models following the two-tower structure cost 7 days to complete the pre-training, and those using the pipeline structure take 3.5 days.

\subsection{Finetuning Details}
\label{subsec:finetune}

We utilize the script\footnote{\url{https://github.com/facebookresearch/SpanBERT/blob/master/code/run_mrqa.py}.} for MRQA in SpanBERT to finetune our models, and the difference are:
the wordpiece-based tokenizer is replaced by the byte-level BPE-based tokenizer, and Huggingface's default optimizer AdamW is used.
We try \texttt{max\_seq\_lenth} 384 and 512.
For the first finetuning setting described in Sec~\ref{subsec:mrqa}, the hyperparameters are selected from learning rates \{1e-5, 2e-5, 3e-5, 5e-5\} and batch sizes \{16, 32\}, and the pre-trained model is finetuned for 4 epochs on a single V100 GPU.
For the second setting, because the training set is large, we directly run experiments using the chosen hyperparameters in \cite{joshi2020contextualized} on 4 V100 GPUs.
The learning rate, batch size and epoch are 2e-5, 32 and 3, respectively.

\subsection{Results on SQuAD 2.0}
\label{subsec:finetune}

\begin{table}[!h]
	\centering
	\begin{tabular}{lcc}
		\hline
		&  \multicolumn{2}{c}{\textbf{SQuAD 2.0}}  \\
		\textbf{Model} & EM & F1   \\
		\hline
		RoBERTa$_{\text{BASE}}$$^{\dag}$ &  80.5 & 83.7 \\
		SpanBERT$_{\text{BASE}}$$^{\dag}$ &  - & 83.6 \\
		ELECTRA$_{\text{BASE}}$$^{\dag}$ &  80.5 & - \\
		\hline
		WKLM$^\ddag$ & 80.4 & 83.4 \\
		KgPLM$_{P}$ (5 epochs) & \bf 81.1 & \bf 84.3 \\ 
		\hline
	\end{tabular}
	\caption{Dev results on SQuAD 2.0. ${\dag}$ denotes that these results are copied from the original paper~\cite{joshi2020contextualized}. $^\ddag$ means the model is re-implemented and trained by us.}\label{tab:mrqa_dev}
\end{table}

\subsection{Comparison with ELECTRA}

This work is indeed inspired by great previous work including ELECTRA, SpanBERT and WKLM. However, there are also significant differences.
ELECTRA is a general pre-training model while our KgPLM specifically targets integrating factual knowledge in the pre-training process. The design of the knowledge masking (K-Mask in Figure~\ref{fig:schemes}) and knowledge replacement (K-Replace in Figure~\ref{fig:schemes}) are key ingredients in the process while not in ELECTRA. Moreover, in ELECTRA, the discriminator is of different size from the generator, while in this work their parameters are shared.

\end{document}